\documentclass{article}

\PassOptionsToPackage{numbers, compress}{natbib}

\usepackage[final]{neurips_2024}

\usepackage[utf8]{inputenc} 
\usepackage[T1]{fontenc}    
\usepackage{hyperref}       
\usepackage{url}            
\usepackage{booktabs}       
\usepackage{amsfonts}       
\usepackage{nicefrac}       
\usepackage{microtype}      
\usepackage{xcolor}         

\usepackage{amsmath}
\usepackage{graphicx}
\usepackage{subfigure}

\title{Long-Range Feedback Spiking Network Captures Dynamic and Static Representations of the Visual Cortex under Movie Stimuli}

\author{%
  Liwei Huang$^{1,2}$, Zhengyu Ma$^{2\ast}$, Liutao Yu$^{2}$, Huihui Zhou$^{2}$, Yonghong Tian$^{1,2,3}$\thanks{Corresponding author.} \\
  \\
  $^{1}$School of Computer Science, Peking University, China \\
  $^{2}$Peng Cheng Laboratory, China \\
  $^{3}$School of Electronic and Computer Engineering, Shenzhen Graduate School, Peking University, China \\
  \texttt{huanglw20@stu.pku.edu.cn,}\\
  \texttt{\{mazhy, yult, zhouhh\}@pcl.ac.cn, yhtian@pku.edu.cn}
}

\begin{document}

\maketitle

\begin{abstract}

Deep neural networks (DNNs) are widely used models for investigating biological visual representations. However, existing DNNs are mostly designed to analyze neural responses to static images, relying on feedforward structures and lacking physiological neuronal mechanisms. There is limited insight into how the visual cortex represents natural movie stimuli that contain context-rich information. To address these problems, this work proposes the long-range feedback spiking network (LoRaFB-SNet), which mimics top-down connections between cortical regions and incorporates spike information processing mechanisms inherent to biological neurons. Taking into account the temporal dependence of representations under movie stimuli, we present Time-Series Representational Similarity Analysis (TSRSA) to measure the similarity between model representations and visual cortical representations of mice. LoRaFB-SNet exhibits the highest level of representational similarity, outperforming other well-known and leading alternatives across various experimental paradigms, especially when representing long movie stimuli. We further conduct experiments to quantify how temporal structures (dynamic information) and static textures (static information) of the movie stimuli influence representational similarity, suggesting that our model benefits from long-range feedback to encode context-dependent representations just like the brain. Altogether, LoRaFB-SNet is highly competent in capturing both dynamic and static representations of the mouse visual cortex and contributes to the understanding of movie processing mechanisms of the visual system. Our codes are available at \href{https://github.com/Grasshlw/SNN-Neural-Similarity-Movie}{\textit{https://github.com/Grasshlw/SNN-Neural-Similarity-Movie}}.

\end{abstract}

\section{Introduction}
\label{sec.intro}

Understanding how the biological visual cortex processes information under natural stimuli with computational models is a critical scientific goal in visual neuroscience. In this realm, deep neural networks have emerged as the predominant tools \cite{khaligh2014deep,kriegeskorte2015deep,yamins2016using}, surpassing traditional models, due to their profound success in matching neural representations \cite{yamins2014performance,cadieu2014deep,cadena2019deep,conwell2021neural,chang2021explaining,nayebi2023mouse}, revealing functional hierarchies \cite{gucclu2015deep,shi2019comparison,dapello2020simulating,vinken2021using,shi2022mousenet}, and explaining functionally specialized processing mechanisms \cite{bakhtiari2021functional,dobs2022brain} of the biological visual cortex. Despite these advancements, research has mainly focused on static image stimuli, leaving a gap in the understanding of neural responses to dynamic, context-rich movie stimuli. This oversight is particularly critical given that the visual system receives predominantly dynamic information and integrates the information in both spatial \cite{hubel1959receptive,hubel1968receptive} and temporal \cite{hasson2008hierarchy} dimensions. To address this challenge, there is a need for models with enhanced biological plausibility, capable of encoding the varied types of information inherent in movie stimuli, in order to deepen our comprehension of the processing mechanisms of the visual cortex.

While bottom-up (feedforward) connections dominate visual processing \cite{felleman1991distributed,dicarlo2012does}, many studies have emphasized the crucial role of top-down (feedback) and lateral connections, which are widespread in the visual cortex \cite{gilbert2013top,kar2019evidence,siegle2021survey}, providing diverse coding mechanisms and augmenting temporal representation protocols \cite{sugase1999global,freiwald2010functional,hupe2001feedback,tang2018recurrent}. This has inspired significant strides in incorporating recurrent structures into computational models, effectively enhancing their ability to emulate brain-like neural representations and biological behavioral patterns \cite{nayebi2018task,kubilius2019brain,kietzmann2019recurrence,rajalingham2022recurrent}. Meanwhile, spiking neural networks (SNNs) \cite{maass1997networks} with brain-like neuronal computational mechanisms have been developed as more biologically plausible models \cite{hodgkin1952quantitative,gerstner2002spiking,izhikevich2004model,brette2007simulation}. Using deep SNNs to model the visual cortex has yielded preliminary success \cite{huang2023deep,zhang2023predicting}. Nonetheless, attempts to combine these biologically plausible structures and mechanisms are lacking.

\begin{figure*}[t]
    \centering
    \subfigure{
    \includegraphics[width=0.95\textwidth]{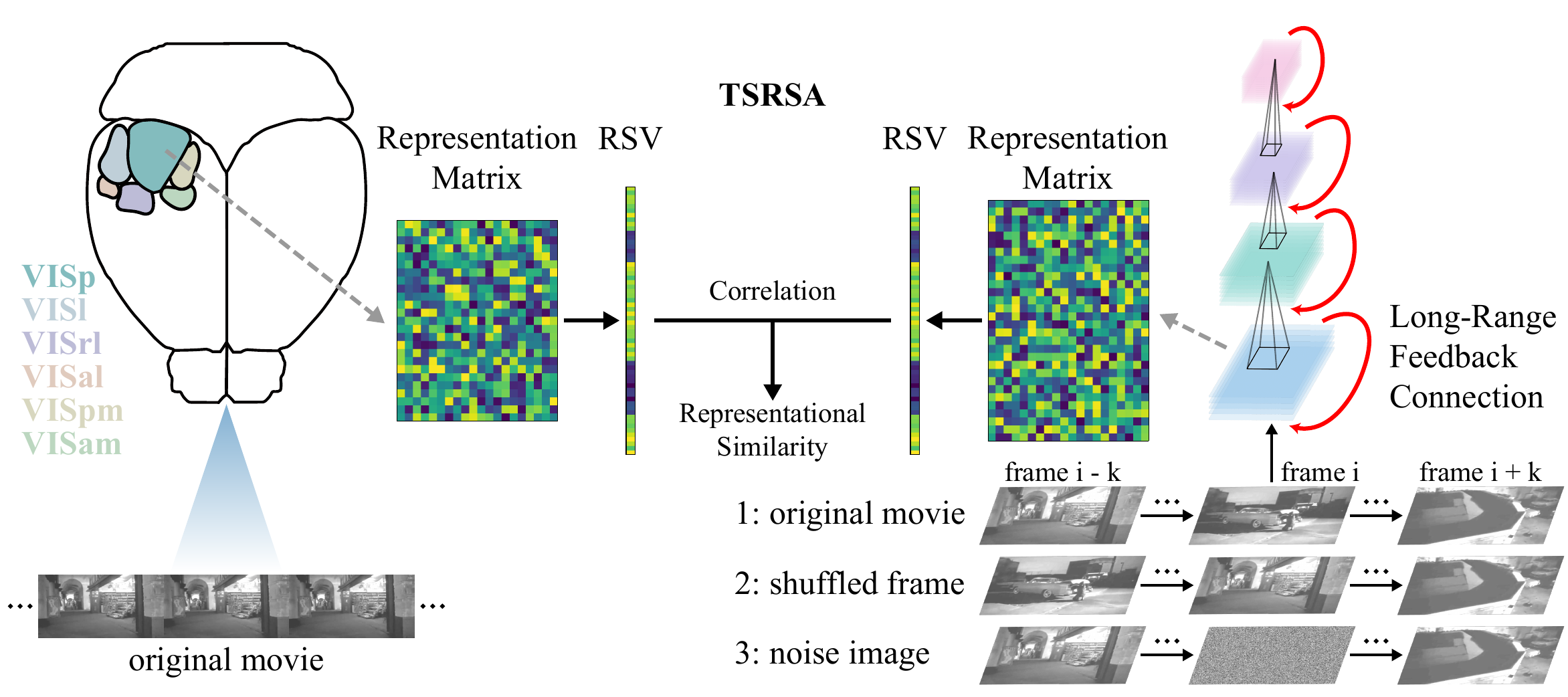}}
    \caption{The overview of our experiments. Six visual cortical regions of the mouse and the long-range feedback spiking network receive the same original movie stimuli to generate the representation matrices. TSRSA is applied to two representation matrices to measure representational similarity. In addition, the network receives two modified versions of the movie stimuli (one with broken temporal structures and the other with varied static textures), while the visual cortex still receives the original movie. These two additional experiments are used to quantify the effects of dynamic (temporal) and static (textural) information on representational similarity. See Section \ref{sec.methods} for details.}
    \label{fig.framework}
\end{figure*}

In this work, we introduce the long-range feedback spiking network (LoRaFB-SNet) to capture both dynamic and static representations of the mouse visual cortex under movie stimuli. To demonstrate the effectiveness of the long-range feedback and the spike mechanism in explaining the information processing mechanisms of the visual system, we design a series of experiments for analyses based on representational similarity (Figure \ref{fig.framework}). The main contributions are as follows.

\begin{itemize}
    \item To mimic top-down connections between cortical regions and to utilize the spike mechanism with dynamic properties, we construct a novel deep spiking network with long-range feedback connections, significantly improving the biological plausibility of the model.
    \item Considering time-dependent sequences of spikes, we propose Time-Series Representational Similarity Analysis (TSRSA) to measure representational similarity between models and the mouse visual cortex.
    \item For neural representations of the mouse visual cortex under movie stimuli, LoRaFB-SNet trained on the UCF101 dataset significantly outperforms other outstanding alternatives in all experiments, demonstrating the critical role of the structures and mechanisms of our model and the training task.
    \item By varying temporal structures or static textures of movie stimuli fed to models, we quantify the effects of dynamic and static information on representational similarity. The results show that LoRaFB-SNet processes movie stimuli to form context-dependent representations in a brain-like manner, providing a deeper insight into the movie processing mechanisms of the visual cortex.
\end{itemize}

Overall, our proposed novel model achieves the highest neural similarity under movie stimuli and better captures brain-like dynamic and static representations, shedding light on the movie coding strategy of the mouse visual cortex.

\section{Related Work}
\label{sec.related}

As deep neural networks have attracted widespread interest as computational models in visual neuroscience, incorporating more biologically plausible structures and mechanisms has become a major avenue to advance the study of neural representations. We summarize some prominent work.

\vspace{-2.5mm}

\paragraph{The deep recurrent network models}

Efforts to construct reasonable recurrent structures \cite{nayebi2018task,kubilius2019brain,kietzmann2019recurrence} have yielded models good at fitting neural representations and revealing neural dynamics. Recurrent cells from an automated search have aided models in predicting the dynamics of neural activity \cite{nayebi2018task}. Notably, CORnet with manually designed architectures has matched the hierarchy of the visual cortex, earning recognition as a leading model in this community \cite{kubilius2019brain}. However, most work has focused on studying neural representations under static stimuli, a limited aspect of stimuli. Besides, while some studies have explored neural responses to movie stimuli \cite{sinz2018stimulus,bakhtiari2021functional,khosla2021cortical}, they have only exploited localized lateral connections.

\vspace{-2.5mm}

\paragraph{The deep spiking network models}

Early studies have applied shallow, even single-layer spiking networks to perform simple temporal tasks \cite{kim2019simple,bellec2020solution,yin2021accurate,rao2022long} and to investigate biological properties \cite{ma2019cortical,suarez2021learning,xue2022spiking}. Some recurrent spiking models have not only emphasized the criticality of homeostatic regulation in biological neurons \cite{ma2019cortical}, but also explained the dynamic regime in the brain associated with cognition and working memory \cite{xue2022spiking}. Recently, deep spiking networks have begun to be used to analyze neural representations, demonstrating enhanced representational similarity \cite{huang2023deep} and more accurate prediction of the temporal-dynamic trajectories of cortical activity \cite{zhang2023predicting} on various neural datasets compared to traditional network counterparts. However, these spiking models are pure feedforward networks and are confined to the study of static stimuli.

\section{Methods}
\label{sec.methods}

\subsection{Long-Range Feedback Spiking Network}
\label{sec.model}

\subsubsection{Architecture}

\begin{figure*}[t]
    \centering
    \subfigure{
    \includegraphics[width=0.8\textwidth]{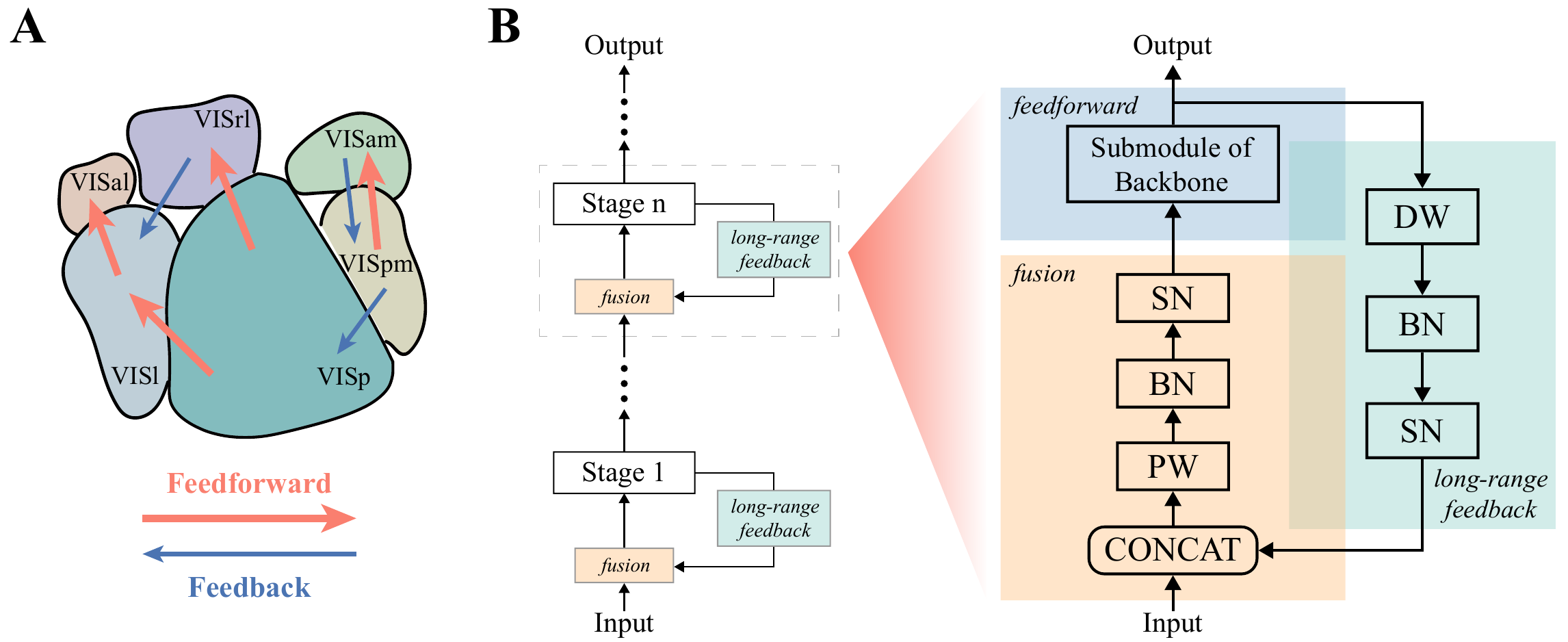}}
    \caption{\textbf{A}. The schematic of six visual cortical regions in the mouse. For brevity, we show parts of the cross-regional feedforward and feedback connections reported from physiological research. \textbf{B}. The schematic of LoRaFB-SNet with the embedded recurrent module. See Section \ref{sec.model} for details.}
    \label{fig.recurrent_module}
\end{figure*}

We develop LoRaFB-SNet guided by two principles of biological plausibility. First, we use the LIF neuron (see Appendix \ref{appendix.spiking_neuron}) as the basic unit of our network, which models the membrane potential dynamics of biological neurons \cite{wu2019direct} and encodes information through spike sequences like the visual cortex. Second, we design the long-range feedback structure to mimic cross-regional top-down connections that are widespread in the mouse visual cortex (Figure \ref{fig.recurrent_module}A). This long-range recurrence is complementary to spiking neurons with self-accumulation for representing temporal information.

As feedforward connections are dominant in the visual cortex, the backbone of LoRaFB-SNet is still feedforward, with the recurrent module embedded to introduce long-range feedback (Figure \ref{fig.recurrent_module}B). In particular, the construction of the recurrent module is described as follows.

The recurrent module consists of three components: a \emph{feedforward} module, a \emph{long-range feedback} module, and a \emph{fusion} module. The \emph{feedforward} module is a submodule of the backbone network, consisting of a stack of convolution, pooling, batch normalization, and spiking neurons, which plays a major role in abstracting spatial features from visual stimuli and encoding the visual content. This module receives the fused features of the outputs from the \emph{long-range feedback} module and the previous stage. The \emph{long-range feedback} module is composed of depthwise transposed convolution, batch normalization, and spiking neurons. On the one hand, depthwise transposed convolution effectively reduces the number of network parameters and upsamples the feature map to match the inputs. On the other hand, some work has shown that such a structure might mimic parallel information processing streams in mouse cortical regions and improve representational similarity \cite{huang2023deep}. The \emph{fusion} module first concatenates the inputs of the current module (the outputs of the previous stage) and the outputs of the \emph{long-range feedback} module in the channel dimension, and then integrates the feedforward and feedback information through pointwise convolution, batch normalization, and spiking neurons. The recurrent module can be formulated as:

\begin{align}
    R^l_t &= {\rm SN}({\rm BN}({\rm DW}(O^l_{t-1}))), \\
    A^l_t &= {\rm SN}({\rm BN}({\rm PW}({\rm CONCAT}(O^{l-1}_t,R^l_t)))), \\
    O^l_t &= {\rm F}^l(A^l_t),
\end{align}

where \({\rm SN}\) is spiking neurons, \({\rm BN}\) is batch normalization, \({\rm DW}\) is depthwise transposed convolution, \({\rm PW}\) is pointwise convolution, \({\rm CONCAT}\) is channel-wise concatenation. \(O^l_t\) denotes the outputs of stage \(l\) at time step \(t\). Similarly, \(R^l_t\) and \(A^l_t\) denote the outputs of the \emph{long-range feedback} and \emph{fusion} modules respectively. \({\rm F}^l\) denotes all operations in the \emph{feedforward} module (Appendix \ref{appendix.feedforward}).

\subsubsection{Pre-Training and Representation Extraction}

We pre-train LoRaFB-SNet on the UCF101 dataset and the ImageNet dataset using SpikingJelly \cite{fang2023spikingjelly}. Specifically, for training the video action recognition task on UCF101, each sample (a video clip) contains 16 frames, and one frame is the input at each time step (the simulating time steps \(T=16\)). For training the object recognition task on ImageNet, each sample (an image) is input to networks 4 times (the simulating time steps \(T=4\)). See Appendix \ref{appendix.train} for details.

After pre-training the networks, we feed them with the same movie stimuli used in the neural dataset and obtain features from all selected layers. For networks trained on UCF101, the entire movie is continuously and uninterruptedly fed into networks in the form of one frame per time step. For networks trained on ImageNet, all frames in the movie are considered as independent images and are fed into networks separately. Each movie frame is input 4 times, which is consistent with training.

\subsection{Representational Similarity Metric}

To assess representational similarity between models and the mouse visual cortex at the population level under temporal sequential stimuli, two main problems need to be addressed. First, the metric can not only analyze static properties of representations, but also preserve temporal relationships of time-series representations to facilitate the analysis of dynamic properties. Second, neurons recorded from the visual cortex are far fewer than units in a network layer, making it difficult to directly compare representations between the two systems.

We present Time-Series Representational Similarity Analysis (TSRSA) based on Representational Similarity Analysis (RSA) \cite{kriegeskorte2008matching,kriegeskorte2008representational}, which has been widely used for the comparison of neural representations \cite{kietzmann2019recurrence,mehrer2020individual,shi2019comparison,bakhtiari2021functional,conwell2021neural}. The original RSA focuses on the similarity between neural representations corresponding to each pair of independent stimuli, whereas TSRSA quantifies the similarity between representations corresponding to sequential stimuli, taking into account temporal sequential relationships. We detail the implementation of TSRSA as follows. First, we acquire representation matrices \(\mathbf{R}=\left(\mathbf{r}_1,\mathbf{r}_2,\ldots,\mathbf{r}_t,\ldots,\mathbf{r}_T\right)\in\mathbb{R}^{{N}\times{T}}\) from each layer of networks and each cortical region, where \(N\) is the number of units/neurons and \(T\) is the number of movie frames. The columns are arranged in chronological order, i.e. \(\mathbf{r}_t\) represents population responses to the movie frame \(t\). Second, we use the Pearson correlation coefficient to compute the similarity between each given column \(\mathbf{r}_t\) and all subsequent columns, yielding the representational similarity vector \(\mathbf{s}_t=\left(s_{t1},s_{t2},\ldots,s_{tp},\ldots\right)\). The element \(s_{tp}\) is \(\mathrm{Corr}\left(\mathbf{r}_t, \mathbf{r}_{t+p}\right)\), where \(0<p<T-t\). We then concatenate all vectors to obtain the complete representational similarity vectors \(\mathbf{S}_\mathrm{model}\) for a network layer and \(\mathbf{S}_\mathrm{cortex}\) for a cortical region, which extract both static features and temporal relationships of neural representations. Finally,  we compute the Spearman rank correlation coefficient between \(\mathbf{S}_\mathrm{model}\) and \(\mathbf{S}_\mathrm{cortex}\) to quantify the similarity. Using this metric, we perform a layer-by-layer measurement for a network, evaluating all selected layers to visual cortical regions. Notably, when obtaining similarity vectors, we choose the Pearson correlation coefficient for computational efficiency, since both model features and neural data are very high-dimensional. On the other hand, the Spearman rank correlation coefficient is chosen to quantify the similarity between two visual systems due to its ability to better capture nonlinear relations.

\subsection{Quantifying Effects of Dynamic and Static Information on Representational Similarity}
\label{sec.quantification}

To analyze how visual models process diverse types of information in movie stimuli, we modify temporal structures (dynamic) and static textures (static) of the original movie and obtain variant dynamic or static representations of networks. Meanwhile, cortical representations are maintained since the movie presented to mice is unchanged. By measuring the similarity between the modified network outputs and the unaltered cortical representations, we quantify the effects of dynamic and static information on representational similarity and attempt to glimpse movie processing mechanisms of the visual cortex from the encoding properties of our model. The methods for modifying temporal structures and static textures are as follows.

\vspace{-2.5mm}

\paragraph{Dynamic information} We disrupt the frame order of the movie and feed the shuffled movie into networks, producing network representations that differ from the original due to distinct dynamic sequential information. To obtain frame order with different levels of alteration while avoiding extreme chaos (e.g., moving the first frame to the last), we divide the entire movie into multiple windows with the same number of frames and randomly shuffle the frames only within each window. We conduct 10 sets of experiments with different window sizes. Each set comprises 10 trials to provide enough statistical power. We calculate the \emph{level of chaos} for every trial, which is defined as \(1-r\), where \(r\) is the Spearman rank correlation coefficient between the disrupted frame order and the original order. Since the movie presented to mice is invariant, we rearrange the network representation matrix to the original order to ensure that it matches the order of the mouse representation matrix when conducting TSRSA. In this way, we maintain the correspondence between static representations of two systems, while isolating changes in dynamic representations of networks. This allows us to focus on evaluating the effects of dynamic information.

\vspace{-2.5mm}

\paragraph{Static information} We randomly select a certain proportion of movie frames and replace them with Gaussian noise images whose static textures are completely different from both network training scenes and biological experiment stimuli. We then feed the new movie into networks to obtain variant static representations. To minimize dense local replacement and preserve as much dynamic information as possible, the movie is divided into equal-sized windows and only one frame in each window is replaced. We similarly run 10 experimental sets with different window sizes, each consisting of 10 trials. The \emph{ratio of replacement} is the inverse of the number of frames per window. Replacing movie frames results in a change in static representations of networks, while the overall frame order remains the same as the original. Admittedly, changing static information will inevitably change temporal structures as well. We attenuate this influence by distributing noise images as sporadically as possible and emphasize how static information affects representational similarity.

\section{Experiments}
\label{sec.experiments}

\subsection{Neural Dataset}
\label{sec.dataset}

We conduct analyses using a subset of the Allen Brain Observatory Visual Coding dataset \cite{de2020large,siegle2021survey}. This dataset, recorded by Neuropixel probes, consists of neural spikes with high temporal resolution from 6 mouse visual cortical regions (VISp, VISl, VISrl, VISal, VISpm, VISam, see Appendix \ref{appendix.dataset} for details). Each region contains hundreds of recorded neurons to minimize the effects of neuronal variability, facilitating the analysis of neural population representations. The visual stimuli presented to the mice consist of two movies, one for 30s (\emph{Movie1}), repeated for 20 trials, and the other for 120s (\emph{Movie2}), repeated for 10 trials. The frame rate of both movies is 30Hz. To pre-process neural responses with the peristimulus time histogram (PSTH), we sum the number of spikes in each movie frame and take the average over all trials for each neuron. To focus on neurons that are more responsive to visual input, we excluded those firing less than an empirical threshold of 0.5 spikes/second \cite{pinto2013fast,williams2021spatial}.

As discussed in the study \cite{de2020large} proposing this dataset, the class of neurons responsive to movie stimuli is found in all six cortical regions. Therefore, for a given model, we take the maximum scores across layers per region and report the average score over six regions as the model's TSRSA score.

\begin{figure*}[t]
    \centering
    \subfigure{
    \includegraphics[width=0.99\textwidth]{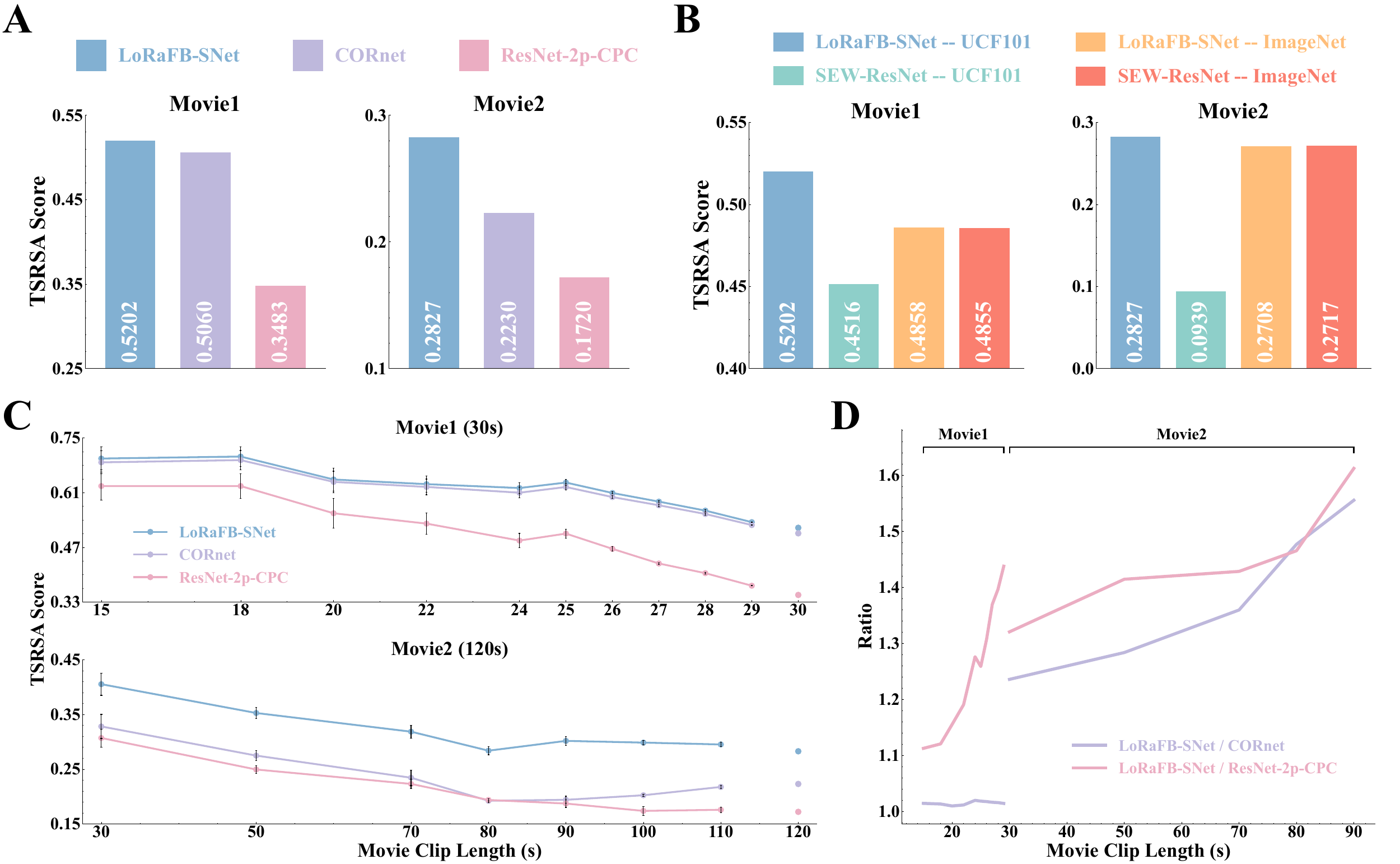}}
    \caption{\textbf{A}. The TSRSA scores of three models pre-trained on UCF101. \textbf{B}. The TSRSA scores of feedback/feedforward spiking networks pre-trained on UCF101/ImageNet. \textbf{C}. The TSRSA score curves of three models pre-trained on UCF101 for different movie clip lengths. We randomly select continuous movie clips of different lengths and plot TSRSA scores between models' and the visual cortex's representations corresponding to these clips. The error bar is the standard error over 10 random seeds. \textbf{D}. The ratios of our model's scores to those of the alternative models for different clip lengths. The ratio tends to increase for longer clips.}
    \label{fig.results1}
\end{figure*}

\subsection{Models for Comparisons}

We select three models for comprehensive comparisons to demonstrate the effectiveness of each character in LoRaFB-SNet.

\vspace{-2.5mm}

\paragraph{CORnet}

It is one of the most influential recurrent networks \cite{kubilius2019brain} for modeling the visual cortex and has been used as a benchmark in many studies \cite{chang2021explaining,conwell2021neural}. The prototype was trained on ImageNet and here we pre-train it on UCF101 with the same training procedure as LoRaFB-SNet. The comparison with CORnet, which has recurrent connections but lacks the spike mechanism, aims to show the critical role of spiking neurons in our model.

\vspace{-2.5mm}

\paragraph{ResNet-2p-CPC}

This model emulates the ventral and dorsal pathways of the mouse visual cortex with parallel pathway architectures \cite{bakhtiari2021functional} pre-trained on UCF101. While our model incorporates spiking neurons and recurrent connections to handle sequential inputs of any length, ResNet-2p-CPC uses fixed-length filters to process temporal data, allowing comparison between different approaches to dynamic information processing.

\vspace{-2.5mm}

\paragraph{SEW-ResNet}

It is a pure feedforward spiking network \cite{fang2021deep} that has shown the best performance in fitting neural representations of the visual cortex \cite{huang2023deep,zhang2023predicting}. We pre-train it on both UCF101 and ImageNet. Since the comparison with it focuses on the role of feedback connections, our model adopts identical feedforward structures to its.

\subsection{Results}
\label{sec.results}

\subsubsection{Comparisons of Representational Similarity}

We perform representational similarity analyses on the neural dataset under two movie stimuli respectively, and refer to the two cases as \emph{Movie1} and \emph{Movie2}.

As shown in Figure \ref{fig.results1}A, among the three models pre-trained on UCF101, LoRaFB-SNet outperforms the other two well-known bio-inspired models. Specifically, our model performs moderately better than CORnet (\emph{Movie1}: \(+2.8\%\); \emph{Movie2}: \(+26.8\%\)) and significantly better than ResNet-2p-CPC (\emph{Movie1}: \(+49.3\%\); \emph{Movie2}: \(+64.4\%\)). To further quantify how similar our model's representations are to brain representations, we obtain neural ceilings by randomly splitting the neural data into two halves and computing the TSRSA score (Table \ref{table.ceiling}). Our model attains \(63.3\%\) and \(45.9\%\) of the ceilings and achieves a great improvement over other models, which suggests that our model effectively captures neural representations of the brain and is meaningfully closer to the mouse visual cortex. We also report similarity scores of our model to each cortical region (Appendix \ref{appendix.region_score}) and show that our model yields robustness across different regions. In addition to the population representation analysis, we use linear regression to fit model representations to temporal profiles of individual biological neurons and compute \(R^2\) as the similarity, the results of which also demonstrate the superiority of our models (Appendix \ref{appendix.regression}).

To emphasize the joint role of long-range feedback connections and pre-training on a video dataset, we compare feedback/feedforward spiking networks pre-trained on image/video datasets (Figure \ref{fig.results1}B). For models trained on ImageNet, LoRaFB-SNet and SEW-ResNet achieve comparable TSRSA scores, suggesting that feedback connections do not have a significant effect on models trained on static images. For models trained on UCF101, LoRaFB-SNet performs significantly better than SEW-ResNet (\emph{Movie1}: \(+15.2\%\); \emph{Movie2}: \(+201.1\%\)). Besides, when comparing models trained on UCF101 and ImageNet, we find that our model outperforms those trained on ImageNet, while SEW-ResNet instead performs worse. This may be explained by the fact that SEW-ResNet trained on a video dataset not only fails to capture dynamic information effectively, but also is compromised on the ability to represent static information. Taken together, it is the combination of long-range feedback connections and pre-training on a video dataset that enables our model to better extract temporal features and capture representations of the visual cortex under movie stimuli.

\begin{table*}[t]
    \caption{The neural ceilings and the scores of all models under two movie stimuli. In brackets are the percentages of model scores compared to the neural ceiling.}
    \begin{center}
    \resizebox{0.99\textwidth}{!}{
        \renewcommand{\arraystretch}{1.2}
        \begin{tabular}{c|c|cccc}
            \hline
            & Neural Ceiling & CORnet & ResNet-2p-CPC & SEW-ResNet & LoRaFB-SNet \\
            \hline
            \textit{Movie1} & 0.821\(\pm\)0.006 & 0.506 (61.6\%) & 0.348 (42.4\%) & 0.452 (55.0\%) & \textbf{0.520 (63.3\%)} \\
            \textit{Movie2} & 0.616\(\pm\)0.009 & 0.223 (36.2\%) & 0.172 (27.9\%) & 0.094 (15.2\%) & \textbf{0.283 (45.9\%)} \\
            \hline
        \end{tabular}
    }
    \end{center}
    \label{table.ceiling}
\end{table*}

Notably, from the above results, we find that our model gains more pronounced advantages in similarity when the movie stimuli are longer. To further corroborate this phenomenon, we randomly select movie clips of different lengths from the original movie stimuli and compute TSRSA scores between models' and the visual cortex's representations corresponding to these movie clips. As shown in Figure \ref{fig.results1}C, TSRSA scores show a decreasing trend with increasing clip length, suggesting that it is more difficult for models to capture brain-like representations when movie stimuli are longer. This result is reasonable since longer movie stimuli increase the diversity of neural response patterns in the visual cortex. Nonetheless, our model consistently outperforms the other two models across all clip lengths and shows increasing improvement ratios as movie clips get longer (Figure \ref{fig.results1}D). These results suggest that the more biologically plausible structure and mechanism may allow LoRaFB-SNet to efficiently process accumulated visual information on longer time scales, just like the brain. Specifically, compared to CORnet which also has recurrent connections, LoRaFB-SNet utilizes the spike firing mechanism to form spike-sequential coding and incorporates long-range feedback to enhance the ability to extract temporal features. Compared to ResNet-2p-CPC, LoRaFB-SNet exploits membrane potential dynamics and recurrent modules to process dynamic information, more flexibly handling movies of different lengths without being limited by the temporal filter size.

Overall, our model, LoRaFB-SNet, outperforms other influential and outstanding alternatives across multiple experimental paradigms, suggesting that biologically plausible long-range feedback connections and spiking neurons significantly contribute to better modeling neural representations of the mouse visual cortex, especially when processing long-duration movies.

\subsubsection{Experiments to Analyze the Effects of Dynamic and Static Information}
\label{sec.information_results}

\begin{figure*}[t]
    \centering
    \subfigure{
    \includegraphics[width=0.99\textwidth]{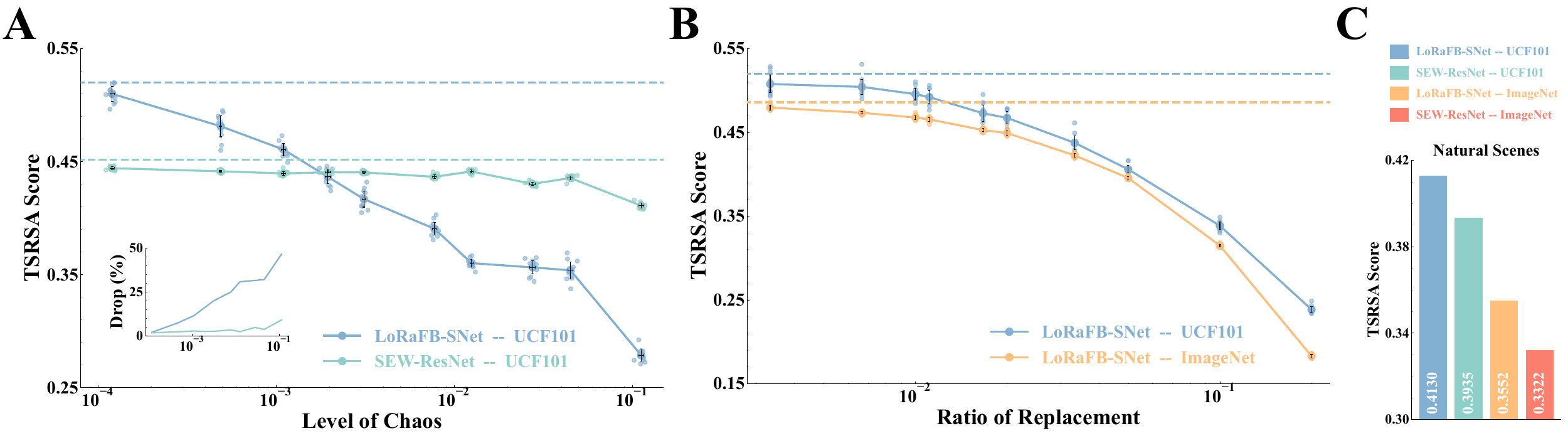}}
    \caption{\textbf{A}. The TSRSA score curves of LoRaFB-SNet and SEW-ResNet trained on UCF101 with different levels of chaos (the main plot) and the drop rate curves of experimental scores compared with the original score (the subplot). The horizontal coordinates in both plots are the level of chaos. In the main plot, the dashed horizontal lines indicate the original scores between models and the mouse visual cortex under the original movie. Each large point on the curve indicates the average result of a set of experiments, and each small point indicates the result of one trial in a set. The vertical error bar is the \(99\%\) confidence interval of the score over 10 trials, while the horizontal error bar is the \(99\%\) confidence interval of the level of chaos. In the subplot, the curves show the average drop rate and the average level of chaos over 10 trials for all experimental sets. LoRaFB-SNet shows a large drop in scores while SEW-ResNet shows a small drop. \textbf{B}. The TSRSA score curves of LoRaFB-SNet trained on UCF101/ImageNet with different ratios of replacement. The elements in the main plot indicate similar content as in \textbf{A}. LoRaFB-SNet trained on UCF101 and ImageNet both exhibit a similar decreasing trend in scores. \textbf{C}. The TSRSA scores of feedback/feedforward spiking networks trained on UCF101/ImageNet for the neural dataset under natural scene stimuli.}
    \label{fig.results2}
\end{figure*}

For the experiments of shuffling movie frames to change dynamic information, we compare the results of LoRaFB-SNet and SEW-ResNet trained on UCF101 (Figure \ref{fig.results2}A) to investigate the importance of brain-like dynamic representations for modeling the visual cortex. As the curves in the main plot show, any alteration from the original frame order results in a lower TSRSA score between models and the mouse visual cortex than the original score, and the score decreases as the level of chaos increases. Frame shuffling disrupts the continuity and temporal structures of the movie, leading to changes in dynamic representations of models. Considering that we align representation matrices of models and the visual cortex along the movie frame dimension when applying TSRSA, the decrease in similarity is mostly attributed to variations in dynamic representations, while the effect of static representations is negligible.

Furthermore, by comparing drop rate curves between LoRaFB-SNet and SEW-ResNet, we find out that the drop rate of LoRaFB-SNet increases with the level of chaos and eventually reaches a staggering \(46.5\%\). However, the drop rate of SEW-ResNet is consistently lower than that of LoRaFB-SNet and the maximum is even lower than \(9\%\). These results reveal two significant findings. First, our model captures biological dynamic representations very well and is sensitive to disruptions in dynamic information. Second, although SEW-ResNet is also trained on a video dataset, the vast majority of its representations depend on static rather than dynamic information. Therefore, its similarity score is less affected by temporal structure distortions. In addition, we perform the same experiment on CORnet that also has recurrent connections (Figure \ref{fig.result4_appendix}A of Appendix \ref{appendix.dynamic_static}), which shows a similar result to our model. In conclusion, these results underscore the superior capability of LoRaFB-SNet, due to its long-range feedback connections, to capture temporal relationships and represent dynamic information in a more context-dependent manner, which is likely to be the potentially crucial mechanism for processing movie stimuli in the mouse visual cortex.

In addition to the analysis of dynamic representations, the role of static representations in modeling the visual cortex should not be overlooked. In the experiments of replacing frames to modify static information, we compare the results of LoRaFB-SNet trained on UCF101 and ImageNet (Figure \ref{fig.results2}B). Similar to the results in the experiments of shuffling movie frames, TSRSA scores are mostly lower than the original in the case of replacing noise frames, and the score decreases as the ratio of replacement increases. We also use other types of noise images for replacement, which exhibits a similar impact (Figure \ref{fig.result4_appendix}B of Appendix \ref{appendix.dynamic_static}). Obviously, static representations of models change a lot due to the totally different static textures between the original movie frames and the noise images, resulting in a decrease in similarity. While scores of both LoRaFB-SNet trained on UCF101 and ImageNet show a similar decreasing trend with the increasing replacement rate, the former is steadily higher than the latter. For the model trained on an image dataset, the movie frames are treated as independent individuals, so model representations completely depend on static information and there is no temporal relationship between representations of two frames. In contrast, the model trained on a video dataset encodes both static and dynamic information to form representations. Consequently, although the high replacement ratio also affects static representations of the UCF101-trained LoRaFB-SNet, its dynamic representations to original movie frames may moderate the drop in similarity score to some extent.

In addition to natural movie stimuli, we measure the representational similarity between models and the mouse visual cortex under static natural scene stimuli (the neural data are also from the Allen Brain Observatory Visual Coding dataset). As shown in Figure \ref{fig.results2}C and Table \ref{table.static_scenes_appendix} of Appendix \ref{appendix.dynamic_static}, our model also outperforms other models in encoding static information and yields more brain-like representations under static stimuli. Besides, models trained on UCF101 perform better than those trained on ImageNet, even in representing static natural scene stimuli.

In summary, LoRaFB-SNet trained on a video dataset is able to extract spatio-temporal features simultaneously and represent dynamic and static information of movie stimuli in a way more similar to the mouse visual cortex. In particular, the powerful ability to encode temporal relationships makes LoRaFB-SNet's representations more context-dependent, which may provide new insights into the mechanisms of movie information processing in the visual system.

\subsubsection{Ablation Studies for Training Datasets and Model Structures}

\begin{table*}[t]
    \caption{The TSRSA scores in different cases for ablation studies under \textit{Movie1}.}
    \begin{center}
        \renewcommand{\arraystretch}{1.2}
        \begin{tabular}{c|ccc|ccc}
            \hline
            & Continuous & Discontinuous & ImageNet & No-spike & SEW-ResNet & LoRaFB-SNet \\
            \hline
            Score & 0.524 & 0.474 & 0.486 & 0.498 & 0.452 & 0.520 \\
            \hline
        \end{tabular}
    \end{center}
    \label{table.ablation}
\end{table*}

To figure out whether the static content of UCF101 data or the temporal structure of continuous videos benefits LoRaFB-SNet to capture brain-like representations, we build two datasets based on UCF101 to pre-train LoRaFB-SNet and compare their similarity. One dataset consists of continuous videos from UCF101, while the other consists of disordered and discontinuous videos made up of randomly selected frames from UCF101. As shown in Table \ref{table.ablation}, LoRaFB-SNet trained on continuous videos outperforms that trained on discontinuous videos, suggesting that the temporal structure rather than the static content plays an important role. Furthermore, LoRaFB-SNet trained on discontinuous videos performs even worse than that trained on ImageNet, showing that the static content of UCF101 is not closer to the test movie stimuli than that of ImageNet in terms of data distribution.

To consolidate the conclusion about the effectiveness of spiking neurons and long-range feedback connections, we perform direct comparisons with two models trained on UCF101, one without spiking neurons but with the same structure as LoRaFB-SNet (No-spike), and the second with the same spiking neurons and feed-forward structure but without feedback connections (SEW-ResNet). The results in Table \ref{table.ablation} show that the model lacking either of these characters performs worse than our model, providing further evidence that both characters of our model play a critical role.

\section{Discussion}
\label{sec.discuss}

In this work, we propose LoRaFB-SNet (Long-Range Feedback Spiking Network) to model neural representations of the mouse visual cortex under movie stimuli. Incorporating long-range feedback connections and spiking neurons, LoRaFB-SNet offers more biologically plausible architectures and processing mechanisms. Tested on the mouse neural dataset with TSRSA, LoRaFB-SNet significantly surpasses existing outstanding and influential computational models across multiple experimental paradigms. We extend the analysis to dynamic and static information representations of networks and the visual cortex with two meticulously designed experiments, providing evidence that LoRaFB-SNet is able to encode dynamic and static information in a more brain-like manner. Specifically, our model efficiently processes visual stimuli with long duration and forms context-dependent representations. Overall, LoRaFB-SNet effectively captures dynamic and static representations of the visual cortex and helps to reveal movie processing mechanisms in the visual system.

For spiking neurons in our model, we hypothesize that membrane potential dynamics and spike information encoding are helpful to better capture brain-like representations, which may require further analyses to support. Benefiting from feedback connections, our model gains particular advantages in representing dynamic information. Some studies \cite{semedo2022feedforward} have suggested that the effects of recurrent connections in the visual cortex vary over time, influencing dynamic representations. The specific contributions and interplay of feedforward and feedback connections to the encoding protocol remain to be explored. In conclusion, while our model explains some mechanisms of information processing in the visual cortex, more biologically plausible mechanisms, such as local recurrent connections and sophisticated neuronal models, deserve to be introduced and studied.

LoRaFB-SNet demonstrates significant efficacy in modeling visual representations of mice. As a biologically plausible spiking network, it holds potential as a general and promising framework for studying the visual cortex of other species and investigating other sensory modalities, helping to understand more intricate neural computations.

\begin{ack}

This work is supported by grants from the Beijing Science and Technology Plan: Z (No. 241100004224011), the National Natural Science Foundation of China (No. 62027804, No. 62425101, No. 62332002, No. 62088102, No. 62206141, and No. 62236009), and the major key project of the Peng Cheng Laboratory (PCL2021A13).

\end{ack}

\bibliographystyle{plain}
\bibliography{ref}

\newpage

\appendix

\section{Spiking Neuron Model}
\label{appendix.spiking_neuron}

The spiking neuron model we used in LoRaFB-SNet is the Leaky Integrate-and-Fire (LIF) model. As mentioned in \cite{fang2021incorporating,fang2021deep}, \(V_t\), \(X_t\) and \(S_t\) denote the state (membrane voltage), input (current) and output (spike) of the spiking neuron model respectively at time step \(t\), and the dynamics of the LIF model can be described as follows:

\begin{align}
    H_t&=V_{t - 1}+\frac{1}{\tau}(X_t-(V_{t - 1}-V_{reset})), \\
    S_t&=\Theta(H_t-V_{thresh}), \\
    V_t&=H_t(1 - S_t)+V_{reset}S_t.
\end{align}

While \(V_t\) is the membrane voltage after the trigger of a spike, \(H_t\) is also the membrane voltage, but after charging and before a spike firing. \(\tau\) is the membrane time constant to control the rate of spiking neuron leakage. \(\Theta(x)\) is the unit step function, so \(S_t\) equals 1 if \(H_t\) is greater than or equal to the threshold voltage \(V_{thresh}\) and 0 otherwise. Meanwhile, \(V_t\) is reset to \(V_{reset}\) when a spike fires. Here, we set \(\tau=2\), \(V_{thresh} = 1\), and \(V_{reset} = 0\), which are widely used empirical values for the visual task training.

Considering that $\Theta(x)$ is non-differentiable at 0, we use the inverse tangent function as the surrogate gradient function \cite{neftci2019surrogate} to approximate the derivative function during back-propagation.

\section{Detailed Structure of Feedforward Module}
\label{appendix.feedforward}

\begin{figure*}[ht]
    \centering
    \subfigure{
    \includegraphics[width=0.99\textwidth]{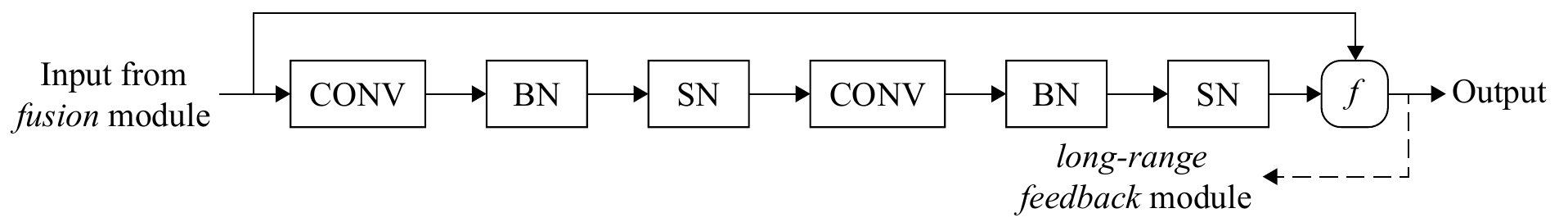}}
    \caption{Detailed structure of the \emph{feedforward} module. \({\rm CONV}\) is convolution. \({\rm BN}\) is batch normalization. \({\rm SN}\) is spiking neurons. \({\textit{f}}\) denotes an element-wise operation with two spike features.}
    \label{fig.feedforward_module_appendix}
\end{figure*}

The \emph{feedforward} module is a submodule of the backbone network, which is made up of a stack of convolution, pooling, batch normalization, and spiking neurons. We adopt the residual block in SEW-ResNet \cite{fang2021deep}, which cures the vanishing/exploding gradient problems of spiking networks. The \emph{feedforward} modules of all stages in LoRaFB-SNet share this structure but with different hyperparameters (Figure \ref{fig.feedforward_module_appendix}).

\section{Pre-Training Implementation}
\label{appendix.train}

In order for networks to extract meaningful features from the visual input, we pre-train them with two visual tasks. Specifically, LoRaFB-SNet, SEW-ResNet and CORnet are pre-trained with the video action recognition task on UCF101, and the first two are also pre-trained with the object recognition task on ImageNet. Notably, ResNet-2p-CPC has already been pre-trained on UCF101 in the original paper \cite{bakhtiari2021functional} and we directly adopt its open-source parameters. The pre-training implementation is detailed as follows.

\paragraph{Pre-Training on UCF101}

In the pre-training procedure on UCF101, all video frames are resized to \(224 \times 224\) and each sample is a video clip of 16 frames that are continuously fed into networks. Since the input at each time step is a video frame, the simulating time steps \(T\) of networks are 16. All networks are trained on UCF101 for 100 epochs on 8 GPUs (NVIDIA V100) with a mini-batch size of 32. The optimizer is SGD with a momentum of 0.9 and a weight decay of 0.0001. The initial learning rate is 0.1 and we apply a linear warm-up for 10 epochs. We decay the learning rate with cosine annealing, where the maximum number of iterations is equal to the number of epochs.

\paragraph{Pre-Training on ImageNet}

In the pre-training procedure on ImageNet, each image is resized to \(224 \times 224\) and fed into spiking networks 4 times. In other words, for each sample, spiking networks are simulated with time steps \(T=4\), and the input is the same image at each time step. We train all spiking networks on ImageNet for 320 epochs on 8 GPUs (NVIDIA V100) with a mini-batch size of 32. We also use SGD as the optimizer and set the momentum to 0.9 and the weight decay to 0. The initial learning rate is 0.1 with a linear warm-up for 5 epochs. Cosine annealing with the maximum number of iterations equal to the number of epochs is also applied to decay the learning rate.

\section{Supplementary Information of the Neural Dataset}
\label{appendix.dataset}

In this work, we use a subset of the Allen Brain Observatory Visual Coding dataset \cite{de2020large,siegle2021survey} recorded from six visual cortical regions of the mouse with Neuropixel probes. The full names and abbreviations of all cortical regions are listed in Table \ref{table.neural_dataset_appendix}. Besides, we present the number of neurons before and after the exclusion of those firing less than 0.5 spikes/s under two movie stimuli. The exclusion criteria resulted in the removal of no more than \(10\%\) of neurons from each region, suggesting that most neurons are responsive.

\begin{table*}[ht]
    \caption{Detailed information of the neural dataset.}
    \begin{center}
        \begin{tabular}{cccc}
        \toprule
        \textbf{Cortical Region} & \textbf{Abbreviation} & \textbf{Total Neurons} & \textbf{Neurons after Exclusion} \\
        &&& \textbf{\textit{(Movie1/Movie2)}} \\
        \midrule
        primary visual cortex & VISp & 2015 & 1880/1854 \\
        lateromedial area & VISl & 933 & 861/851 \\
        rostrolateral area & VISrl & 1415 & 1302/1267 \\
        anterolateral area & VISal & 1553 & 1445/1420 \\
        posteromedial area & VISpm & 879 & 820/807 \\
        anteromedial area & VISam & 1506 & 1394/1366 \\
        \bottomrule
        \end{tabular}
    \end{center}
    \label{table.neural_dataset_appendix}
\end{table*}

\section{TSRSA Scores of LoRaFB-SNet to Each Cortical Region}
\label{appendix.region_score}

We report TSRSA scores of our model to each cortical region of the mouse. The results show that our model achieves stable scores across regions and there is no significant difference.

\begin{table*}[ht]
    \caption{The scores of LoRaFB-SNet to each cortical region under two movie stimuli.}
    \begin{center}
        \renewcommand{\arraystretch}{1.1}
        \begin{tabular}{c|cccccc}
            \hline
            & VISp & VISl & VISrl & VISal & VISpm & VISam \\
            \hline
            \textit{Movie1} & 0.527 & 0.501 & 0.505 & 0.515 & 0.530 & 0.544 \\
            \textit{Movie2} & 0.222 & 0.262 & 0.296 & 0.300 & 0.301 & 0.315 \\
            \hline
        \end{tabular}
    \end{center}
    \label{table.region_score_appendix}
\end{table*}

\section{Results of Linear Regression for Individual Neurons}
\label{appendix.regression}

\begin{figure*}[ht]
    \centering
    \subfigure{
    \includegraphics[width=0.99\textwidth]{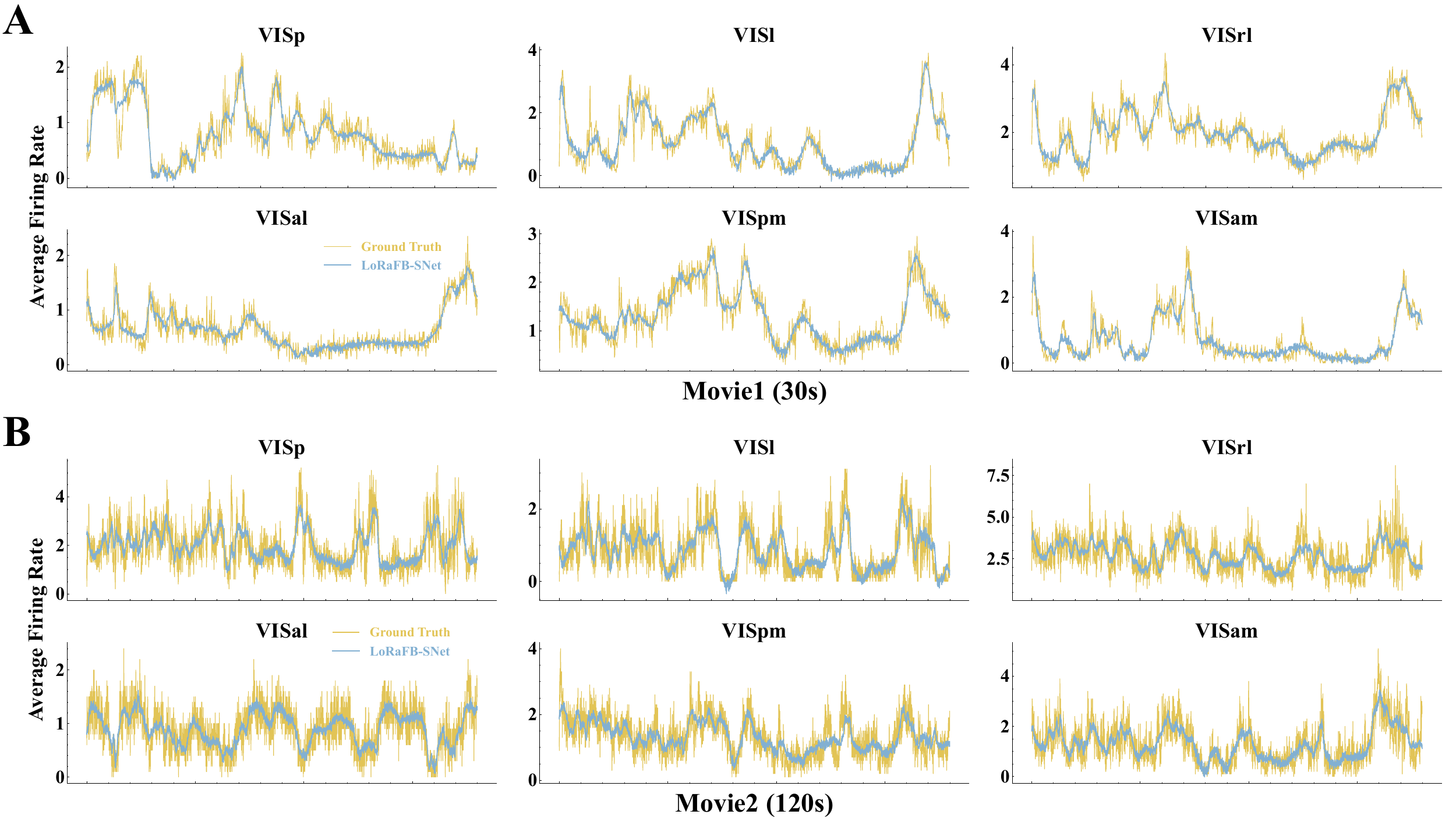}}
    \caption{The real and regressed temporal profiles of individual biological neurons. We choose one neuron in each cortical region as an example.}
    \label{fig.results3_appendix}
\end{figure*}

We use linear regression to fit model representations to temporal profiles of individual neurons and report \(R^2\) as the similarity. The results show that our model consistently performs better than alternative models on this metric (Table \ref{table.regress_appendix}). In addition, we present some examples of real temporal profiles of biological neurons and the regressed results of our model (Figure \ref{fig.results3_appendix}), which also demonstrate the good fitting performance of our model.

\begin{table*}[ht]
    \caption{The similarity scores of all models using the regression-based metric.}
    \begin{center}
        \renewcommand{\arraystretch}{1.1}
        \begin{tabular}{c|cccc}
            \hline
            & CORnet & ResNet-2p-CPC & SEW-ResNet & LoRaFB-SNet \\
            \hline
            \textit{Movie1} & 0.4326 & 0.4167 & 0.4241 & \textbf{0.4335} \\
            \textit{Movie2} & 0.1790 & 0.1751 & 0.1720 & \textbf{0.1836} \\
            \hline
        \end{tabular}
    \end{center}
    \label{table.regress_appendix}
\end{table*}

\section{Additional Results of Experiments for Dynamic and Static Information}
\label{appendix.dynamic_static}

\begin{figure*}[ht]
    \centering
    \subfigure{
    \includegraphics[width=0.99\textwidth]{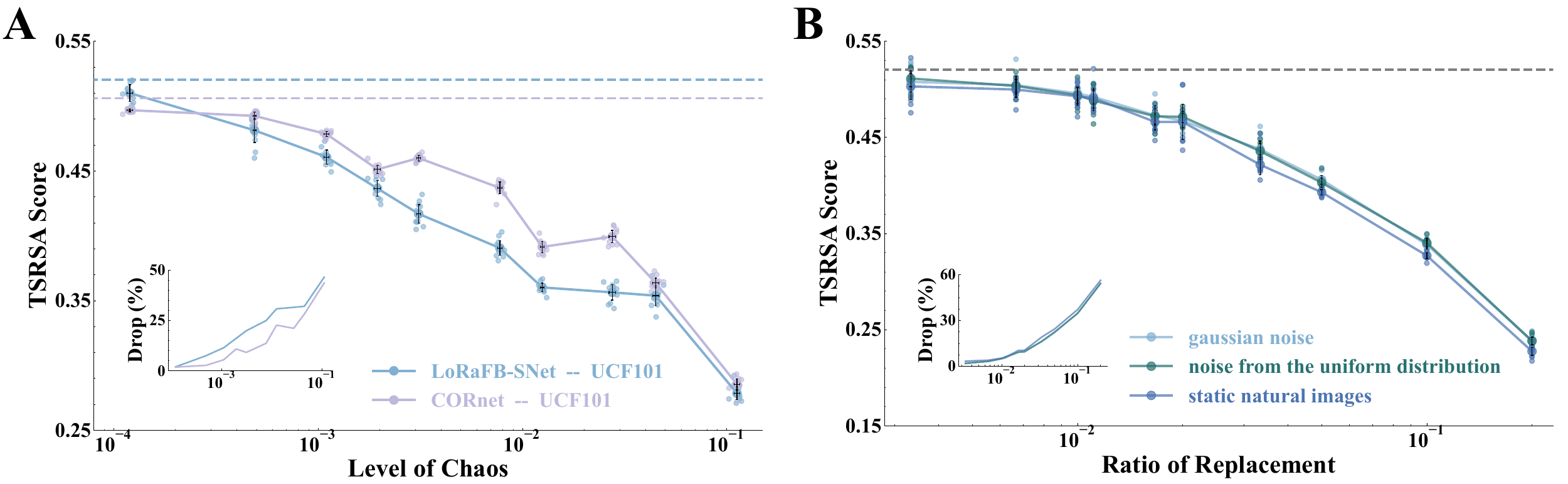}}
    \caption{\textbf{A}. The TSRSA score curves of LoRaFB-SNet and CORnet trained on UCF101 with different levels of chaos. The elements in the plot indicate similar content as in Figure \ref{fig.results2}A. \textbf{B}. The TSRSA score curves of LoRaFB-SNet trained on UCF101 with different ratios of replacement. There are three types of noise images. The elements in the plot indicate similar content as in Figure \ref{fig.results2}B.}
    \label{fig.result4_appendix}
\end{figure*}

For the experiments of changing dynamic information, CORnet yields a similar result to our model (Figure \ref{fig.result4_appendix}A), which solidifies our conclusion about the effectiveness of feedback connections. For the experiments of modifying static information, we use two other types of images for replacing frame experiments, including noise images from the uniform distribution and static natural images from the Allen Brain Observatory Visual Coding dataset. The results of UCF101-trained LoRaFB-SNet are shown in Figure \ref{fig.result4_appendix}B, suggesting that various noise images for replacement all have a similar impact on representational similarity. Besides, we report similarity scores of all models under static natural scenes stimuli in Table \ref{table.static_scenes_appendix} and show that our model achieves the highest score.

\begin{table*}[ht]
    \caption{The similarity scores of all models under static natural scenes stimuli.}
    \begin{center}
        \renewcommand{\arraystretch}{1.1}
        \begin{tabular}{c|cccc}
            \hline
            & CORnet & ResNet-2p-CPC & SEW-ResNet & LoRaFB-SNet \\
            \hline
            \textit{Static Scene} & 0.3544 & 0.3435 & 0.3935 & \textbf{0.4130} \\
            \hline
        \end{tabular}
    \end{center}
    \label{table.static_scenes_appendix}
\end{table*}

\end{document}